\documentclass[letterpaper, 10 pt, conference]{ieeeconf}
\IEEEoverridecommandlockouts
                                      \usepackage{epsfig}
\usepackage{amssymb}
\usepackage{amsmath}
\usepackage{cite}
\usepackage{booktabs}
\usepackage{multirow}
\usepackage{mathtools}
\usepackage{bm}
\usepackage{algorithm}
\usepackage{comment}
\usepackage{color}
\usepackage{setspace}
\usepackage{tablefootnote}
\usepackage{makecell}
\usepackage{tikz}
\usepackage{gensymb}
\usepackage[noend]{algorithmic}

\title{\LARGE \bf Adaptive Grasping of Moving Objects in Dense Clutter \\ via Global-to-Local Detection and Static-to-Dynamic Planning}

\author{Hao Chen, Takuya Kiyokawa, Weiwei Wan, and Kensuke Harada 
\thanks{*All authors are with the Department of Systems Innovation, Graduate School of Engineering Science, Osaka University, Toyonaka, Osaka 560-8531, Japan
{\tt\small h.chen@hlab.sys.es.osaka-u.ac.jp} and {\tt\small\{kiyokawa, wan, harada\}@sys.es.osaka-u.ac.jp}}%
}

\begin{document}
\maketitle
\thispagestyle{empty}
\pagestyle{empty}

\begin{abstract}
Robotic grasping is facing a variety of real-world uncertainties caused by non-static object states, unknown object properties, and cluttered object arrangements. The difficulty of grasping increases with the presence of more uncertainties, where commonly used learning-based approaches struggle to perform consistently across varying conditions. In this study, we integrate the idea of similarity matching to tackle the challenge of grasping novel objects that are simultaneously in motion and densely cluttered using a single RGBD camera, where multiple uncertainties coexist. We achieve this by shifting visual detection from global to local states and operating grasp planning from static to dynamic scenes. Notably, we introduce optimization methods to enhance planning efficiency for this~time-sensitive task. Our proposed system can adapt to various object types, arrangements and movement speeds without the need for extensive training, as demonstrated by real-world experiments.
\end{abstract}

\section{INTRODUCTION}
In logistic warehouses, a wide variety of daily items are transported on conveyor belts every day, and human workers are required to pick out target items from an unorganized clutter and pack them into delivery boxes. To achieve robotic automation for this task, an interesting issue is posed: grasping unknown objects in clutter moving on a conveyor belt.

Existing studies on robotic grasping for unknown objects predict grasps directly from partial point clouds \cite{c2,c3} or perform shape completion before planning \cite{c4,c5}. However, they have limited generalization to object types outside of their training dataset due to their data dependency. Moreover, these methods mainly focus on single or static target objects, whereas in real-world scenarios, objects are often in cluttered or dynamic states where multiple uncertainties coexist.

Therefore in this study, we aim at developing a grasping strategy that can cope with unknown object types and high-uncertainty scenes where the objects are moving in dense clutter. The key to achieving this challenging task is the proposed global-to-local detection and static-to-dynamic planning framework. During visual detection, we first capture the object features used for grasp planning from a global image, then assess their movement states from a local viewpoint. For grasp planning, we first perform static planning to generate robust grasp poses, followed by dynamic planning to enable real-time grasping of the objects in motion. The combination of global-to-local detection and static-to-dynamic planning provides an effective solution for grasping moving objects in clutter that are previously unseen, since it addresses multiple real-world uncertainties step-by-step rather than all at once as in traditional learning-based methods. The experimental results demonstrate the exceptional effectiveness of our method in grasping novel objects under significant uncertainties.

Our main contributions can be summarized as:

\begin{itemize}
\item We propose a global-to-local visual detection and static-to-dynamic grasp planning approach to robustly grasp moving objects in dense clutter on a conveyor belt. 

\item We develop an adaptive object tracking method that~accurately estimates the movement state of object clutters.

\item We develop a recurrent algorithm to enable continuous dynamic grasping of moving objects in dense clutter.
\end{itemize}

\begin{figure}[t]
    \centering
    \includegraphics[width=0.9\linewidth]{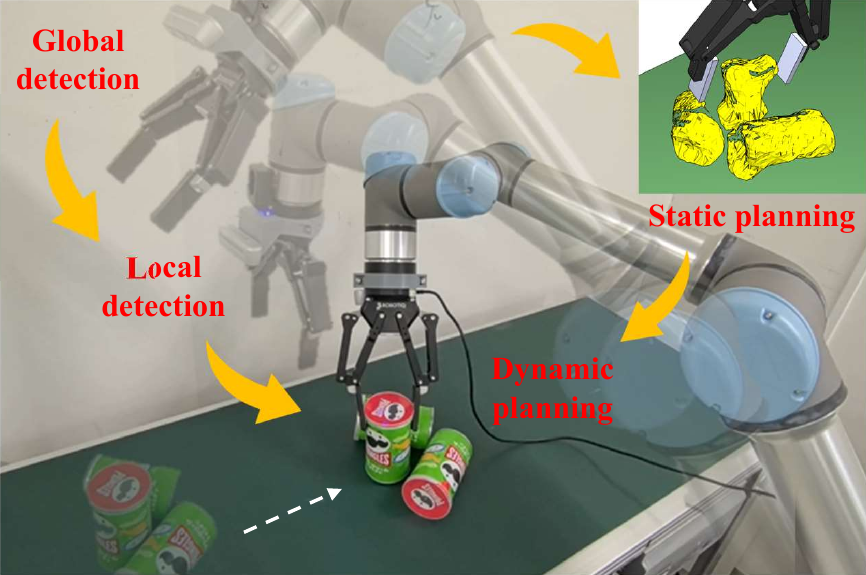}
    \caption{Grasping moving cluttered cans by our method. Static and dynamic planning are executed at the end of global and local detection respectively.}
    \label{f1}
\end{figure}

Fig. \ref{f1} showcases the general flow of our system. We utilize the flexibility of an in-hand camera to switch between global and local observation poses, and leverage the results of static grasp planning to achieve real-time dynamic object grasping.

\begin{figure*}[t]
    \centering
    \includegraphics[width=0.9\linewidth]{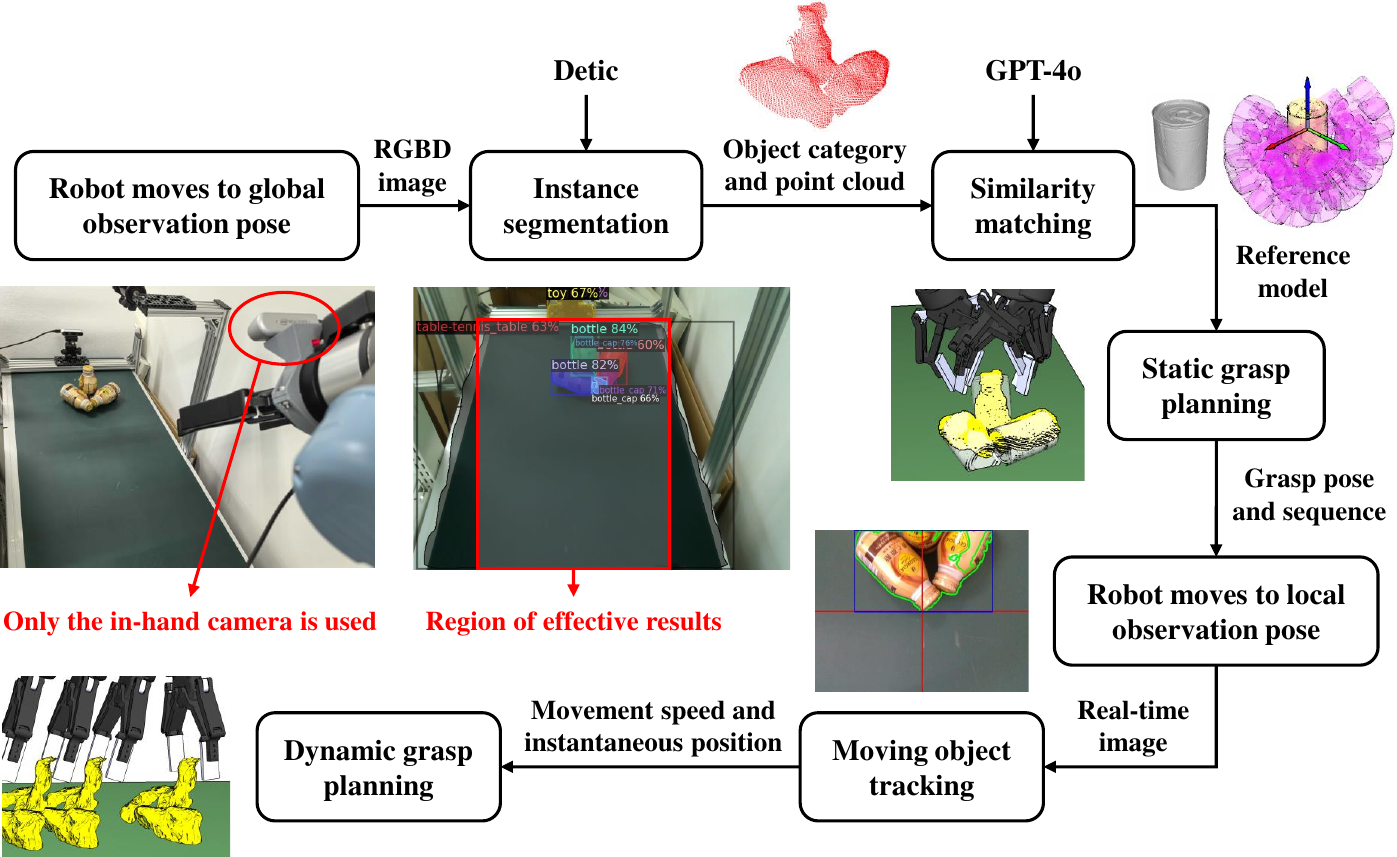}
    \caption{Workflow of our proposed system. All visual detections and grasp planning are performed online using only a single in-hand camera.}
    \label{f2}
\end{figure*}

\section{RELATED WORKS}
Recent research on grasping unknown objects typically addresses cluttered and moving objects separately, but few can combine these two types of uncertainty due to the greatly increased difficulty in grasp detection and model training.

\textbf{Grasping objects in clutter.} As a representative of this research topic, GraspNet-1Billion \cite{c7} is a state-of-the-art generalized object grasping model trained on a large-scale dataset consisting of a large number of RGBD images and grasp poses. HGGD \cite{c8} is a more recent work that proposes a local grasp generator combined with grasp heatmaps as guidance to efficiently detect real-time grasps for cluttered objects. In addition to training directly on RGBD images or point clouds, VGN \cite{c9} and GIGA \cite{c10} propose to use Truncated Signed Distance Fields (TSDFs) to represent cluttered object scenes and utilize this representation to learn grasp detection. There are also reinforcement learning (RL) based approaches, such as QT-Opt \cite{c11} and VPG \cite{c12}, which collect large amounts of training data in a self-supervised manner and utilize skills such as pushing to achieve high success rates for cluttered object grasping. These methods share common drawbacks such as high-cost, time-consuming training process, and limited adaptability to varying scenes. In contrast, our method is training-free and does not require specific operating environments.

\textbf{Grasping objects in motion.} Dynamic grasping of moving objects has recently attracted attention in the field of robotic grasping. GG-CNN \cite{c13} is a well-known generative grasping network that is capable of handling dynamic scenes where object positions are changed after each grasp attempt. However, they need the objects to remain stationary during the grasp execution. Marturi et al. \cite{c14} achieve adaptive grasping for various types of moving objects by developing a local planner for object tracking and a global planner for grasp switching. However, they require prior observation of the target object from multiple viewpoints to obtain its complete surface geometry. Two recent works \cite{c15,c16} use Recurrent Neural Network (RNN) and Long Short-Term Memory (LSTM), respectively, to predict the future locations of moving objects for precise dynamic grasping. However, they require the target object to be pre-trained with robust grasps, whereas our method can handle novel objects without any prior knowledge. In addition, a few studies \cite{c17,c18} incorporate RL algorithms to achieve moving object grasping with a single camera. However, their methods have only been evaluated on single objects, not in cluttered scenes.

\section{METHODS}

\subsection{System Overview}
Our goal is to grasp unknown objects from a dense clutter moving on a conveyor belt. Fig. \ref{f2} shows the complete system workflow. Similar to an industrial setting, the cluttered objects are initially positioned at the start of the conveyor belt, with the robot located some distance away. To achieve the grasping task, we first let the robot move to a global observation pose where the in-hand camera can observe the entire clutter from a diagonal downward view. In this pose, we use the camera to capture an instant RGBD image and perform instance segmentation on the RGB input to obtain object categories, and combine the results with the depth input to obtain object point clouds. Based on these two types of information, we implement a similarity matching method similar to \cite{c6} to obtain a similar reference model for each object in the clutter from an existing database. During this process, we utilize LLMs to greatly improve the matching efficiency. Based on the reference models, we perform static grasp planning to simultaneously generate robust grasp poses for all objects in the clutter at a fixed position in front of the robot. The grasp sequence for different objects is also determined by the planning results.

During static grasp planning, an adaptive local observation pose with a top-down view is also generated based on object locations and arrangements. When this pose is determined, we immediately let the robot move there and start capturing real-time images to assess the movement state of the clutter. We incorporate a moving object tracking method to obtain the movement speed and instantaneous position of the clutter based on two key time points. Based on these two results, we finally develop a recurrent algorithm to achieve dynamic grasping for moving cluttered objects.

\subsection{Global Visual Detection} 
We use a single in-hand camera for visual detection for two reasons: 1) For cluttered objects, it is difficult to identify object correspondences in different viewpoints when using multiple cameras; 2) We need a diagonal downward view to get more surface information about the objects, and a top-down view to accurately track their movement, in which case a flexible in-hand camera becomes a better choice. 

During global detection, the camera obtains an instant RGBD image containing the cluttered objects and feeds it into a SOTA instance segmentation model of Detic \cite{c19}. To exclude the segmentation results of background objects, we define a region based on the position of the conveyor belt in the image and extract only the results within that region. For redundant results such as \textit{bottle\_cap} in \textit{bottle}, we ignore them by calculating the containment relationship between detected bounding boxes. From the filtered segmentation results, we can obtain the category name and 3D point cloud of each object in the clutter. Based on them, we can perform similarity matching to find reference models from an existing database to guide the grasping of unknown objects, as was done in \cite{c6}. However, their method requires matching with all database models, which is computationally long and not suitable for the task of grasping moving objects.

\subsection{LLM-Assisted Similarity Matching}
Considering that most database models are irrelevant to the target object, we improve matching efficiency by using LLMs (GPT-4o \cite{c20} in our task) to pre-screen potential candidates based on object categories, passing only these candidates for further matching. The implementation is as follows:

\textbf{Prompt}: Which objects in the \{\textit{YCB dataset}\} are likely to be similar to \{\textit{can}\} in terms of robotic grasping? Please list the 5 most likely object names with their indices.

\textbf{Answer}: ... 005\_tomato\_soup\_can, 006\_mustard\_bottle, 010\_potted\_meat\_can, 021\_bleach\_cleanser, 025\_mug ...

In the prompt, the first bracket can be filled in with an existing model database that is known to LLMs and can be used for similarity matching, and the second bracket can be filled in with the obtained category name of each object in the clutter. From the answer, we can extract only the key information about the selected candidates by recognizing their indices. For these candidates, we further perform point cloud registration between their point clouds and the obtained point cloud of each object in the clutter using the RANSAC \cite{c21} and ICP \cite{c22} algorithms. Due to the pre-screening of model candidates, registration can be finished in a short time.

\begin{figure}[t]
    \centering
    \includegraphics[width=0.95\linewidth]{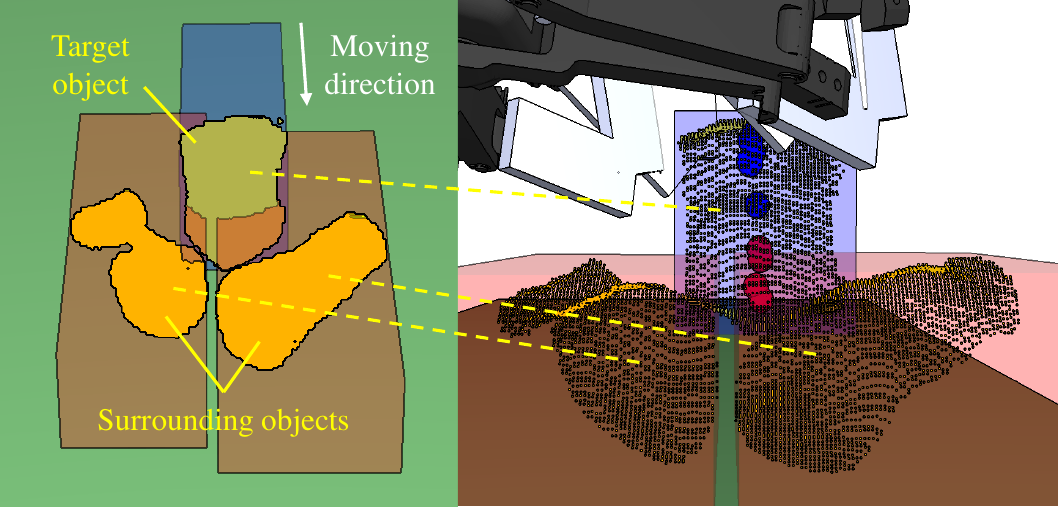}
    \caption{Fast and accurate grasp planning for cluttered objects based on \textit{collision area} (blue $\cup$ red) and \textit{overlap area} (blue $\cap$ red).}
    \label{f3}
\end{figure}

\begin{figure}[t]
    \centering
    \includegraphics[width=0.85\linewidth]{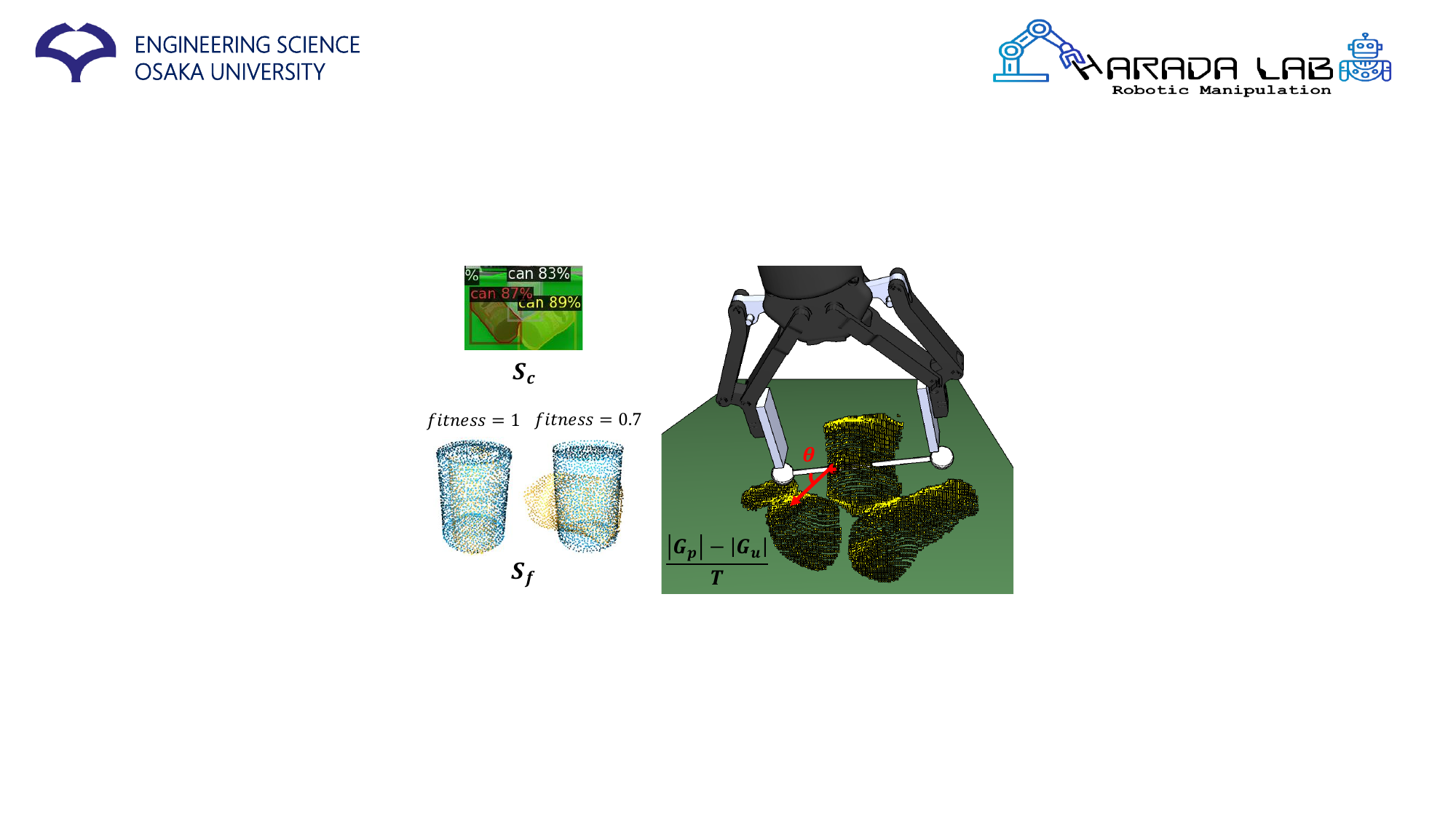}
    \caption{Multi-metric evaluation of grasp sequences in cluttered scenes.}
    \label{f4}
\end{figure}

\begin{figure*}[t]
    \centering
    \includegraphics[width=0.9\linewidth]{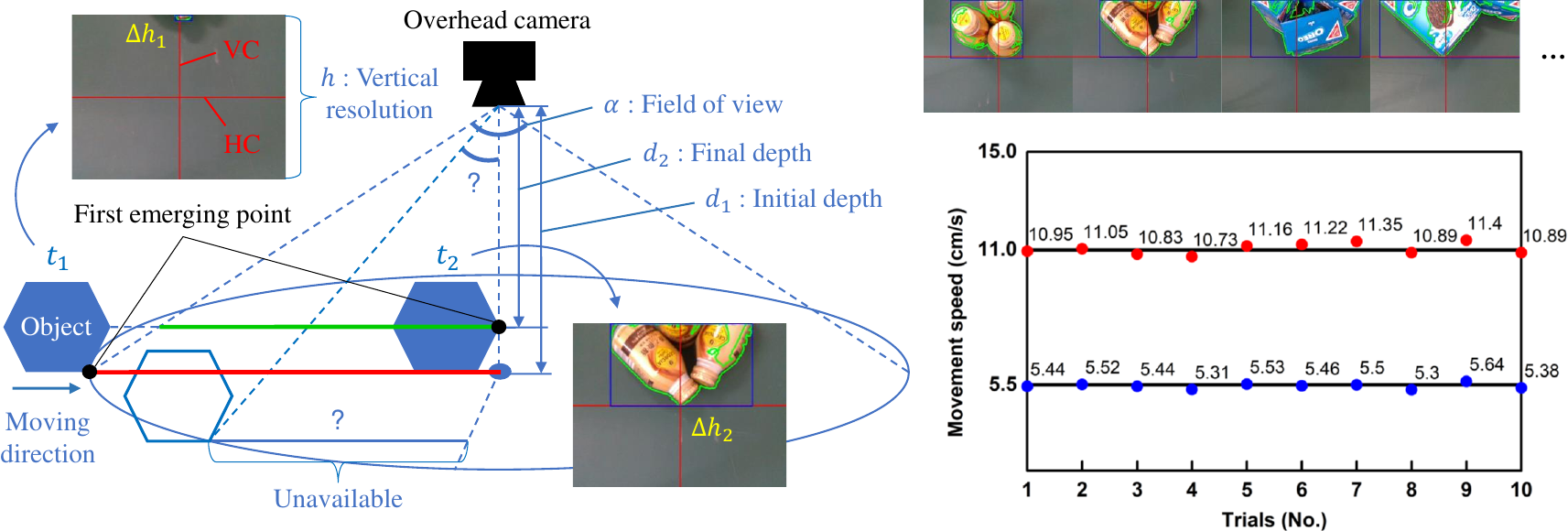}
    \caption{(Left) Speed estimation of moving cluttered objects based on adaptive local visual detection and real-time object tracking. (Right) Evaluation of the calculation results in two conveyor speed modes (5.5 cm/s and 11 cm/s). In each trial, we change either the object arrangement or object type.}
    \label{f5}
\end{figure*}

\subsection{Static Grasp Planning}
Each model in the database is pre-planned with over one hundred antipodal grasps using a mesh surface segmentation approach \cite{c23}. These grasps are robust through an accurate analysis of the complete mesh model. For grasp planning of an unknown object, we transfer the pre-planned grasps from a similar database model to the unknown object based on the transformation matrix obtained from point cloud registration, similar to what was done in \cite{c6}. The grasp planning process is performed in a robot simulation environment \cite{c24}.

In our task, the objects are moving and we need to quickly generate grasps for real-time dynamic grasping. For this purpose, we first assume that the cluttered objects are right in front of the robot and perform static grasp planning at this fixed position $p_0$. Then during dynamic grasping, we directly query the planned static grasp poses to quickly generate real-time grasps. To ensure the efficiency and accuracy of grasp planning, we define two types of areas called \textit{collision area} and \textit{overlap area} based on the axis-aligned bounding box (AABB) of each object in the clutter, as shown in Fig.~\ref{f3}. In each grasp planning, there is one target object and several surrounding objects. For the target object, we extend its AABB in the opposite direction of the moving direction to generate a blue area, and for the surrounding objects, we extend their AABBs in both the positive and negative directions of the moving direction to generate a red area. The concatenation and intersection of the blue area and the red area indicate \textit{collision area} and \textit{overlap area}, respectively. 

To avoid collisions between the gripper and all moving objects before and after grasping, the pregrasp pose defined by backing off the grasp pose for a small distance cannot be located within \textit{collision area}. In addition, at the grasp pose, the target object cannot collide with the gripper fingers and should be located within the gripper's closure area. Based on these rules, we exclude all infeasible grasps included in the grasps transferred from similar models. To further accelerate the planning process, we prioritize the grasps whose centers are outside \textit{overlap area}. As shown in Fig. \ref{f3} right, the grasp poses with blue centers are more likely to be collision-free than those with red centers in cluttered scenes.

Another important issue is the determination of the grasp sequence. While the grasp planning for all objects in the clutter is performed simultaneously, they need to be grasped in a certain order. As shown in Fig. \ref{f4}, we consider several factors for this including: 1) the confidence score during instance segmentation $S_c$, as it reflects the object occlusion rate and the reliability of the similarity matching results; 2) the fitness score during point cloud registration $S_f$, as it determines the quality of the grasps transferred from similar models; and 3) the speed of the grasp generation, as faster grasp planning represents lower grasping difficulty. Both $S_c$ and $S_f$ can be directly obtained from the corresponding algorithms, while the speed of the grasp generation needs to be calculated over a time period $T$ ($T=5s$ in our task). We take the total number of IK-solvable and collision-free potential grasps generated within this time period $\left | G_p \right | $ minus the number of unstable grasps $\left | G_u \right |$ as the evaluation metric. The grasp stability is evaluated based on the relative angle $\theta$ between the gripper opening direction and the estimated normal direction of the contact point. $\theta$ is taken as an acute angle, as shown in Fig. \ref{f4} right. When $\theta>30\degree$ at any contact point, we consider the grasp unstable and categorize it into $G_u$. When contact points are within the invisible area causing $\theta$ to be unavailable, we categorize such grasps as potential grasps and use them only when none of the grasps in $G_p$ are evaluated as stable. Finally, we develop a multi-metric function for determining the grasp sequence as follows:
$$
P=S_c\ast S_f\ast \frac{\left | G_p \right |-\left | G_u \right |}{T} \eqno{(1)}
$$
where $P$ denotes the grasp priority of each object in the clutter. Each time after planning grasps for the object with the highest priority, we remove that object from the grasp planning of the other objects and recalculate the grasp priority for the remaining objects.

\subsection{Adaptive Local Detection}
In a dynamic grasping task, we need to obtain the movement state of moving objects in addition to appropriate grasp poses. For this purpose, we move the in-hand camera to an overhead position that can observe the objects from a top-down view perpendicular to the moving direction. We use a frame-difference based object tracking algorithm \cite{c25} to obtain the 2D bounding box for cluttered objects in real-time images. By utilizing the fact that the camera's field of view (FOV) is known, we can accurately estimate the clutter's movement speed by capturing two key time points: 1) $t_1$, when the clutter just enters the camera view; 2) $t_2$, when the clutter reaches the horizontal centerline (HC) of the camera view. The movement speed can be obtained by dividing the distance the clutter traveled between $t_1$ and $t_2$ by the time difference. However, since the FOV of a typical camera is only known in vertical and horizontal directions, the travel distance is only available when the clutter enters from the vertical centerline (VC) of the camera view, as shown in Fig.~\ref{f5} left. Therefore, we adaptively set the camera position so that the frontmost point of the clutter is aligned with VC, and the camera height is maintained at a certain distance (30 cm in our task) above the highest point of the clutter. This can be easily achieved since the point clouds of all objects in the clutter have been obtained during global visual detection.

Two other factors that significantly affect the accuracy of speed estimation are: 1) the first emerging point, which represents the part of an object that first appears at a certain position in the camera view, as shown in Fig. \ref{f5} left. When this point changes between $t_1$ and $t_2$, the real travel distance is an unknown value between the lengths of the green and red lines. However, we can approximate it by obtaining the depth of the first emerging point at $t_1$ and $t_2$ (denoted as $d_1$ and $d_2$, respectively) and averaging their values; 2) the detection latency, which is an unavoidable error occurring when the moving clutter has traveled a short distance beyond the target line before being detected. This error becomes significant at higher movement speeds and can be compensated by accounting for the number of pixels traversed, denoted as $\Delta h_1$ and $\Delta h_2$ for $t_1$ and $t_2$, respectively. We can then derive the following equation for accurate speed estimation:
$$
\widetilde{v} =\frac{(d_1+d_2)/2 \ast \tan (0.5+(\Delta h_2-\Delta h_1)/h)\alpha}{t_2-t_1} \eqno{(2)}
$$
where $\alpha$ is the camera's FOV in the moving direction and $h$ is the vertical resolution of the camera view. We validate the efficiency of this calculation method by testing in two conveyor speed modes, each with 10 trials using different types or arrangements of cluttered objects, as shown in Fig.~\ref{f5} right. The average test results for the two speed modes are $5.45\pm 0.10$ (cm/s) and $11.05\pm 0.23$ (cm/s), both of which are very close to the ground truth with minimal fluctuation.

In addition, we can also obtain the instantaneous position $p_2$ of the clutter at $t_2$ since the camera position is known, and perform dynamic grasp planning based on $\widetilde{v}$ and $p_2$.

\subsection{Dynamic Grasp Planning}
Using the results of static grasp planning and local visual detection, we develop a recurrent algorithm to achieve dynamic grasping for moving objects in dense clutter (see Algorithm \ref{alg:1}). In the first step of grasping the first-priority object, we set a time interval of $\Delta t=1$s and predict the future position of the clutter at each subsequent $\Delta t$ after $t_2$ based on $\widetilde{v}$ and $p_2$. Taking the future state $t_3$ as an example, we translate the static grasps to the predicted position $p_3$ and check their feasibility. When there is more than one feasible grasps at $p_3$, we select the closest grasp and let the robot move to its pregrasp pose. During this process, we record both the time used for grasp planning and the time used for the robot to move from the local observation pose to the pregrasp pose as $t_p$ and $t_m$, respectively. Meanwhile, we assume the time taken by the robot to approach the grasp pose from the pregrasp pose to be $t_a=0.5$s. If $t_p+t_m+t_a<\Delta t$, it means that the robot has enough time to complete the grasping motion before the clutter reaches $p_3$. In this case, we let the robot wait for a small period of time (the time difference) and then execute the grasp. Otherwise, we proceed to the next future state and repeat the process from planning to moving until the time condition is satisfied.

In the subsequent steps of grasping lower-priority objects, we delay $t_2$ by the amount of time taken to grasp the previous objects and repeat the same process as in the first step. Each step is done in a separate thread to prevent conflicts.

\renewcommand{\algorithmicrequire}{\textbf{Input:}}
\renewcommand{\algorithmicensure}{\textbf{Output:}}
\begin{algorithm}[t]
\caption{Dynamic grasp planning for cluttered objects}\label{alg:1}
\begin{algorithmic}[1]
\REQUIRE Grasp poses and sequences from static planning, estimated movement speed and instantaneous position\\[-\baselineskip]
\STATE \textbf{Step 1:} Grasp the first-priority object $o_1$
\STATE Initialize $n=3$, $t_c=0$s, $t_a=0.5$s and $\Delta t=1$s
\WHILE{True}
\STATE $t_n=t_{n-1}+\Delta t$ (Initially, $t_3=t_2+1$s)
\STATE Predict the clutter's future position $p_n=p_{n-1}+\widetilde{v}\Delta t$\\[-\baselineskip]
\FOR {each static grasp pose $g\in o_1$}
\STATE Translate the grasp from $p_0$ to $p_n$
\STATE Compute IK solutions and check collisions
\IF{$g$ is IK-solvable and collision-free}
\STATE Save $g$ into $G_f$
\ENDIF
\ENDFOR
\IF{$G_f\ne \emptyset$}
\STATE Find the optimal grasp pose $g^*\in G_f$ with the minimum distance to the local observation pose
\STATE Let the robot move to the pregrasp pose of $g^*$
\STATE Record the planning time $t_p$ and motion time $t_m$
\STATE Calculate the cumulative time $t_c=t_c+t_p+t_m$
\IF{$t_c+t_a<(n-2)\Delta t$}
\STATE Wait for $(n-2)\Delta t-(t_c+t_a)$ seconds
\STATE Let the robot move to $g^*$ and execute the grasp
\STATE \textbf{break}
\ENDIF
\ENDIF
\STATE $n=n+1$
\ENDWHILE
\STATE \textbf{Step 2:} Grasp the second-priority object $o_2$
\STATE Record the entire duration of grasping the first object $t_d$\\[-\baselineskip]
\STATE In another thread, delay $t_2$ by $t_d$ with all other parameters unchanged and repeat the same process as in Step~1\\[-\baselineskip]
\STATE \textbf{Step 3 ($o_3$), Step 4 ($o_4$), ... :} Same as Step 2
\end{algorithmic}
\end{algorithm}

\section{EXPERIMENTS}
\begin{figure}[t]
    \centering
    \includegraphics[width=0.9\linewidth]{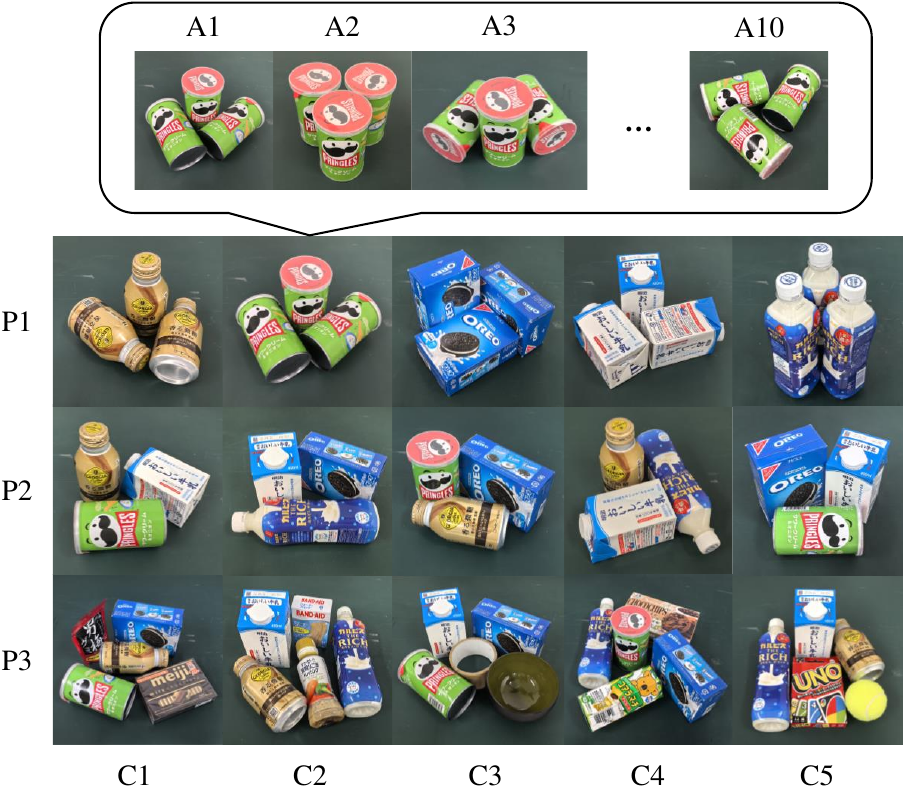}
    \caption{Experimental objects in 3 patterns, 5 clutters, and 10 arrangements.}
    \label{f6}
\end{figure}

\subsection{Experimental Setup}
We carry out grasping experiments using a UR5e robot arm equipped with a Robotiq 2F-140 adaptive gripper, an in-hand RealSense D435 depth camera and a standard-type conveyor belt. The existing database we use for similarity matching is the YCB dataset \cite{c26}, excluding objects without mesh models or with distorted models. All computations are performed on a PC equipped with a Ryzen 7 5800H CPU and a GeForce RTX 3060 GPU. The Detic model and GPT-4o model are pre-loaded to reduce task processing time.

\subsection{Dynamic Grasping Experiments}

\begin{table*}[t]
\small
\renewcommand\arraystretch{1.2}
\setlength\tabcolsep{7pt}
\centering
\caption{Experimental Results on the SR and ER for Grasping Various Types of Moving Clutter}
\begin{tabular}{c | c | c c c c c | c | c c c c c | c}
\toprule 
\multicolumn{2}{c|}{Speed mode} & \multicolumn{6}{c|}{Conveyor speed = 5.5 cm/s} & \multicolumn{6}{c}{Conveyor speed = 11.0 cm/s}\\
\hline
\multicolumn{2}{c|}{Clutter set} & C1 & C2 & C3 & C4 & C5 & Average & C1 & C2 & C3 & C4 & C5 & Average\\ 
\hline 
\multirow{2}{*}{P1} & SR & 76\% & 88\% & 94\% & 82\% & 80\% & 84\% & 67\% & 80\% & 80\% & 78\% & 70\% & \textbf{75\%}\\
& ER & 85\% & 80\% & 95\% & 85\% & 75\% & 84\% & 90\% & 100\% & 100\% & 90\% & 100\% & \textbf{96\%}\\
\hline 
\multirow{2}{*}{P2} & SR & 88\% & 89\% & 82\% & 78\% & 93\% & \textbf{86\%} & 67\% & 70\% & 78\% & 75\% & 63\% & 71\%\\
& ER & 85\% & 90\% & 85\% & 90\% & 75\% & \textbf{85\%} & 90\% & 100\% & 100\% & 80\% & 90\% & 92\%\\
\hline 
\multirow{2}{*}{P3} & SR & 77\% & 73\% & 79\% & 75\% & 88\% & 78\% & 60\% & 63\% & 57\% & 75\% & 67\% & 64\%\\
& ER & 65\% & 75\% & 70\% & 60\% & 85\% & 71\% & 100\% & 80\% & 70\% & 80\% & 90\% & 84\%\\
\bottomrule
\end{tabular}
\label{tab:1}
\end{table*}

We select various types of novel objects and make them into dense clutter for grasping experiments. In order to verify the effectiveness and generalizability of our method, we categorize the clutters into three patterns (see Fig. \ref{f6}): (P1) three identical objects; (P2) three different objects; (P3) five different objects. For each pattern, we create five different clutter sets (C1-C5). For each clutter set, we generate ten different object arrangements (A1-A10). Thus, we conduct experiments on a total of $3\times5\times10=150$ clutter scenarios.

We test in two conveyor speed modes (5.5 cm/s and 11.0 cm/s) and evaluate the results by two metrics: \textbf{Success Rate (SR)} and \textbf{Execution Rate (ER)}, calculated as SR = Number of Successful Grasps / Number of Grasps Performed, ER = Number of Grasps Performed / Number of Objects Targeted. Due to time constraints in the dynamic grasping task, we set the number of objects targeted to two in the 5.5 cm/s speed mode and one in the 11.0 cm/s speed mode. The objects to be grasped are determined autonomously by the planning process. In general, SR and ER can be used to represent the accuracy and efficiency of our method, respectively. 

A complete task cycle is: Place the cluttered objects at the start of the conveyor belt $\to$ Run the conveyor, start visual detection and grasp planning $\to$ Grasp the first object and place it in a nearby box $\to$ Grasp and place the second object (only in the 5.5 cm/s scenario) $\to$ Stop the conveyor. An object is considered successfully grasped if it is placed in the box without being dropped during the process.

The experimental results are shown in Table \ref{tab:1}. Overall, our method effectively handles moving clutter with varying object numbers, types and arrangements, demonstrating both high success rates and execution rates. In both speed modes, P1 and P2 yield similar results, indicating that our method performs consistently with multiple object types coexisting. In the low-speed mode, most execution failures occur when attempting to grasp the second object (as the clutter reaches the end of the conveyor belt before a grasp can be executed), resulting in lower ERs compared to the high-speed mode where only one object is targeted. In contrast, more grasping failures are observed under higher conveyor speeds due to the reduced tolerance for positional error during grasping. When the number of objects increases, as in P3, the processing~time rises and the presence of more surrounding obstacles increases the likelihood of collisions, leading to lower SRs and ERs compared to fewer object scenes such as P1 and~P2. 

In addition, we compare the performance without using LLM in similarity matching and find that the ER drops below 50\% at low speed and nearly 0\% at high speed, highlighting its critical importance in achieving optimal task efficiency.

\subsection{Failure Analysis}
For both grasp failures and execution failures, we summarize the reasons and possible solutions as follows:

1) \textbf{Large occlusion rate.} When the cluttered objects are arranged in such a way that a large part is invisible to the camera, the extracted object point clouds are too sparse for stable similarity matching and grasp planning results. A possible solution is to add more viewpoints or observe more times to obtain more complete appearance of the objects.

2) \textbf{Too dense arrangement.} When the objects are closely packed and there are too few graspable areas, the planning process takes a long time and the clutter may have reached the end of the conveyor belt before a feasible grasping motion is output. A possible solution is to combine other skills, such as pushing objects, to create more space for grasping.

3) \textbf{Dynamic collision.} When the first object is grasped and moved to the placement position, other objects in the clutter are still moving, and collisions may occur between the robot and these moving objects during the pick-and-place process. A possible solution is to develop a real-time motion planning method to avoid collisions in such dynamic scenes.

\section{CONCLUSIONS}
In this study, we present a novel framework using global-to-local detection and static-to-dynamic planning to achieve grasping of moving objects in dense clutter. Especially, we use an improved similarity matching method to efficiently plan grasp poses and sequences for cluttered objects, and propose an adaptive object tracking method to accurately estimate the movement speed and the instantaneous position of the moving clutter. Based on these results, we develop a recurrent algorithm to achieve continuous dynamic grasping.

The effectiveness and generalizability of our method is verified through real-world experiments. However, additional methods can be combined to address the limitations of the current approach in future work.

\addtolength{\textheight}{-9.5cm}   




\section*{ACKNOWLEDGMENT}

This research is subsidized by New Energy and Industrial Technology Development Organization (NEDO) under a project JPNP20016. This paper is one of the achievements of joint research with and is jointly owned copyrighted material of ROBOT Industrial Basic Technology Collaborative Innovation Partnership (ROBOCIP).

\end{document}